\documentclass{ppig}
\usepackage{epsfig}
\usepackage{booktabs}
\usepackage{ucs}
\usepackage[utf8x]{inputenc}

\usepackage{csquotes}
\usepackage{todonotes}
\usepackage{caption}
\usepackage{subcaption}

\title{Story-thinking, computational-thinking, programming and software engineering}

\author{Austen Rainer\\
  School of Electronics, \\
  Electrical Engineering \\
  \& Computer Science\\
  Queen's University Belfast \\
  a.rainer@qub.ac.uk \\
  \And
  Catherine Menon \\
  School of Physics, \\
  Engineering  \\
  \& Computer Science \\
  University of Hertfordshire\\
  c.menon@herts.ac.uk \\
  }

\date{June 2022}

\begin{document}
\maketitle
\thispagestyle{empty}

\begin{abstract}
Working with stories and working with computations require very different modes of thought. We call the first mode ``story-thinking'' and the second ''computational-thinking''. The aim of this curiosity-driven paper is to explore the nature of these two modes of thinking, and to do so in relation to programming, including software engineering as programming-in-the-large. We suggest that story-thinking and computational-thinking may be understood as two ways of attending to the world, and that each both contributes and neglects the world, though in different ways and for different ends. We formulate two fundamental problems, i.e., the problem of ``neglectful representations'' and the problem of oppositional ways of thinking. We briefly suggest two ways in which these problems might be tackled and identify candidate hypotheses about the current state of the world, one assertion about a possible future state, and several research questions for future research.

\end{abstract}

\section{Introduction}
\label{section:introduction}

The term ``story'' is widely used in software engineering, e.g., the ``user story''~\cite{lucassen2015forging} and the ``job story''~\cite{lucassen2018jobs}. Related concepts are also used, e.g., the scenario~\cite{carroll2003making}. But \textit{story} -- in the fullest sense of that word -- and algorithm are very different things. We use the term ``story-thinking'' to refer to the ways in which we conceptualise and engage with stories, e.g., how we create them, how we formally analyse and evaluate them, and how and why we appreciate them as readers.
We use the term ``computational-thinking'' to refer to the ways in which we think about computations, e.g., how we create algorithms, how we formally analyse and evaluate them, and how we simulate their execution on real or nominal machines. The contrasts between story-thinking and computational-thinking stimulate a range of questions, e.g., what are the ways in which story and algorithm might relate? how might story -- it's writing, telling and re-telling -- complement computational thinking? and what do we we gain, and what do we lose, with these contrasting ways of thinking?

The aim of this curiosity-driven paper is to explore the nature of these two modes of thinking, and to do so in relation to programming, including software engineering as programming-in-the-large. The paper contributes the following:
\begin{enumerate}
    \item A conceptualisation of some of the issues, through the application of these two modes of thinking to a six word story.
    \item The identification of two fundamental problems: the problem of ``neglectful representations'' and the problem of oppositional ways of thinking.
    \item Brief proposals for how these problems might be tackled, including prospective hypotheses and research questions.
\end{enumerate}

The remainder of the paper is organised as follows. We start our exploration with a thought experiment in Section~\ref{section:thought-experiment}. In Section~\ref{section:computational-representations-of-story}, we model the thought experiment computationally. In Section~\ref{section:computational-thinking}, we discuss the concept of ``computational-thinking'', addressing what it means to think computationally. In Section~\ref{section:programming-in-the-large}, we then consider computational thinking in the context of software engineering, as programming-in-the-large. This leads into the consideration of requirements engineering, in Section~\ref{section:requirements-engineering}, as a presumed ``bridge'' between the humanly-meaningful real world and the computational world. In Section~\ref{section:modes-of-thinking}, we briefly consider related perspectives on modes of thinking, a neuro-scientific perspective and a psychological perspective. In Section~\ref{section:the-problem}, we then summarise the problems, propose ways forward, suggest two hypotheses, one assertion and several research questions. Finally, with Section~\ref{section:conclusion}, we briefly conclude.




\section{A thought experiment}
\label{section:thought-experiment}

Consider the following short story (which has been intentionally modified slightly from a well-known story in the literary world):
\begin{displayquote}
    ``For sale. Baby’s shoes. Never worn.''
\end{displayquote}

What is your reaction to this story? 

In his book, \textit{Once upon an if}, Worley \citeyear{Worley2014} describes his wife's first reaction to this story. ``Oh, no!'' he says she responded, an indication that she interpreted the story as one of sadness or tragedy, presumably relating to the life of a baby. As the reader, you might, similar to Worley's wife, interpret the story as a tragedy. Or you might, as Worley says his wife's friend did, interpret the story as about someone who buys things for others but those things are not wanted. And, of course, you might have other reactions to the story. In a recent online presentation of this story to software engineering academics, the audience was asked for their reactions. Responses included, ``love, compassion'', ``sadness'', ``cute'', and ``nostalgic -- story made me reflect on that episode of my life''.

Furthermore, the reference to a \textit{baby} means the phrase ``never worn'' carries different connotations compared to, for example, an adult having ``never worn'' the shoes. The connotations of a baby's shoes never being worn are tragic because babies are seen as more susceptible to, and ``fitting'' (in story-terms) for, tragedy. An adult's ``never worn'' shoes may imply the shoes have been worn at least once, for example to try them on, but then not worn again, e.g., perhaps they didn't fit.

Whatever your interpretation of the story, and your reaction to it, the story-teller (based on a story allegedly written by Ernest Hemingway, though this is disputed) has provoked an experience and done so very efficiently.
In just six words, the story-teller creates characters (e.g., a baby, and possibly a parent), a plot (e.g., perhaps someone has lost a baby, or not been able to have a baby) and therefore an unfolding of events over time, one or more goals and a struggle (e.g., the goal of having a baby, with the struggle of not having a baby, or of surviving the loss of a baby), an outcome (e.g., the goal was not attained) and an emotional experience for the reader (e.g., sadness).

The entire story is shorter in length than Cohn's~\citeyear{cohn2004user} well-accepted \textit{template} for a \textit{single} user story in requirements engineering: \texttt{As a <type of user> , I want <goal>, [so that <some reason>]}. One reason we chose this story is because it is concise and therefore easy to present completely in an academic publication with restrictions on page length. But we also chose this story because it is efficient: an extraordinary amount of information and emotion is evoked in only a few words. This efficiency contrasts with the user story. The contrast -- between the six-word story and the template for a single user-story -- suggests fundamental differences in the way that stories model the world and the way that typical software engineering and programming constructs model the world; and also suggests fundamental differences in the ways that story--thinking and computational-thinking require us, or encourage us, to \textit{attend} to the world.

\section{Representing the story computationally}
\label{section:computational-representations-of-story}

An alternative way to think about the six-word story is computationally. And it seems that as soon as we start to attend to the six-word story computationally it is no longer a \textit{story} but instead becomes a text. To illustrate this point we present and consider two forms of computational thinking: user stories and software designs. There are other forms we might consider too, e.g., we might write an algorithm and instantiate it as a program. We don't explicitly consider algorithms or programs here for two reasons. First, due to space; second, because the two forms we do consider here would typically be prerequisites to then developing the algorithm and the program. We do consider the transformation of representations later in this paper when we discuss software engineering.

\subsection{Representing the story as user stories}
\label{subsection:representation-as-user-story}

One way to approach the thought experiment is to represent the story as user stories. Table~\ref{table:user-stories} presents examples of \textit{possible} user stories, first re-stating Cohn's~\citeyear{cohn2004user} template for user stories. These user stories can only be speculative because they depend on how one interprets the six words.

\begin{table}[ht]
\centering
\small
\begin{tabular}{c l } 
    \hline
    ID & Example  \\ 
    \hline\hline
    N/A & As a <type of user> , I want <goal>, [so that <some reason>]  \\ 
    1 & As a grieving mother, I want to sell my baby's shoes, so that I can reduce my financial losses.\\
    2 & As a grieving parent, I want to sell my baby's clothing, so that I can recover my financial losses.\\
    3 & As the purchaser of an unwanted item of clothing, I want to sell the item, so that I can recover my costs.\\
    4 & As a user, I want to be able to sell items, so that I can make some money.\\
    \hline
\end{tabular}
\caption{Example user stories for the thought experiment}
\label{table:user-stories}
\end{table}

The examples in Table~\ref{table:user-stories} illustrate some of the tensions, strengths and weaknesses of computational-thinking and story-thinking, such as:
\begin{enumerate}
    \item The \textit{story} tells us very little explicitly about the actual protagonist, other than the protagonist wants to sell a pair of baby's shoes. We are left to infer the characteristics of the protagonist, including the basis of their goal. In terms of story-thinking, this is an effective rhetorical device. In terms of computational-thinking, this is problematic. There is a tension  between story-thinking's use of evocative connotation and computational-thinking's standards of specification and denotation.
    \item There is also a tension between story-thinking's particularity and computational-thinking's abstraction. The users, or personas, defined in User Stories 1 through 3 can all be abstracted to the user defined in User Story 4. Conversely, whilst the emotional aspects of the users in User Stories 1, 2 and 3 (e.g., sadness, nostalgia, not wanting something) might be \textit{representable} in a software system, it is not clear what gain there is \textit{for the software system} with such a representation. Or in other words, it is not clear how representing such states in the system helps the user with what they want to \textit{do}.
    
    This distinction between the particularity and the generality of the story can be further illustrated by comparing the version of the story presented at the beginning of Section~\ref{section:thought-experiment} with the original story. In the original, the story reads, ``For sale. Baby shoes. Never worn.'' Adding one apostrophe together with one letter, the letter \textit{s}, helps to particularise the story.
    
    
    \item In the act of preparing User Stories to model the story, we begin to reshape our conception of the story. We specify, and possibly also abstract, and by doing so we change the story, limit it and reduce its effect as a story.
\end{enumerate}

\subsection{Representing the story as a software design}
\label{subsection:representation-as-software-designy}

\begin{figure}[ht]
     \centering
    \begin{subfigure}[b]{0.1\textwidth}
         \centering
         \label{}
     \end{subfigure}
     \hfill
     \begin{subfigure}[b]{0.25\textwidth}
         \centering
         \includegraphics[width=\textwidth]{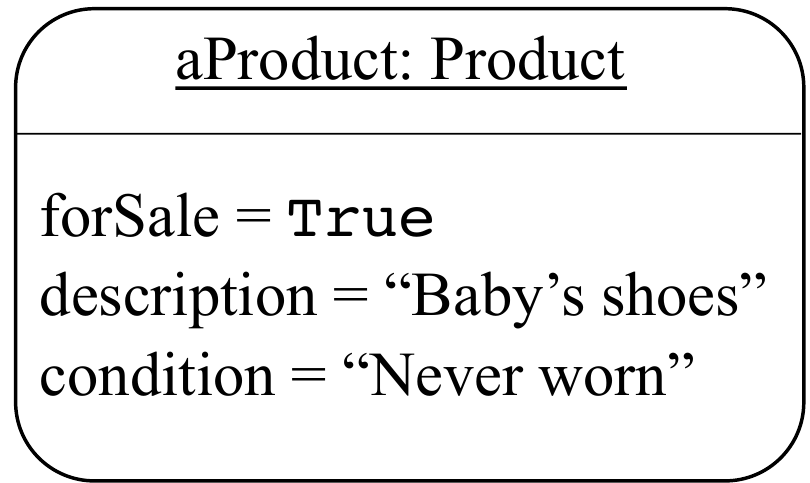}
         \caption{UML object diagram}
         \label{subfigure:for-sale-object-diagram}
     \end{subfigure}
     \hfill
     \begin{subfigure}[b]{0.55\textwidth}
         \centering
         \includegraphics[width=\textwidth]{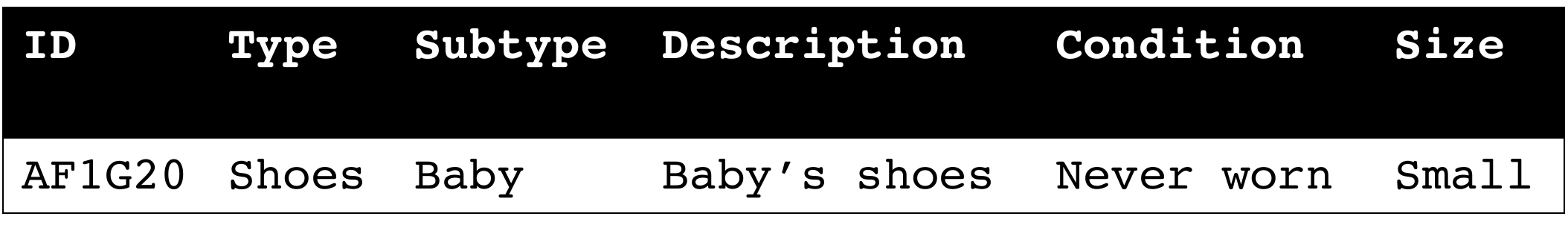}
         \caption{Database table}
         \label{subfigure:for-sale-database-table}
     \end{subfigure}
     \hfill
     \caption{UML object diagram and database table for the sale of a product}
     \label{figure:for-sale-diagrams}
\end{figure}

A second way to think computationally about the story is in terms of software designs. Figure~\ref{figure:for-sale-diagrams} presents a simple UML object diagram (Figure~\ref{subfigure:for-sale-object-diagram}) and a database table (Figure~\ref{subfigure:for-sale-database-table}) for the sale of a product, in this case a pair of baby's shoes. We can of course think with and about these models, but notice how the \textit{kind} of thinking we do with these models is different to the kind of thinking we do with the \textit{story}.

With the object diagram we can, for example, infer that the object is an instance of the class \texttt{Product}, and we can see the \texttt{forSale} attribute is a boolean variable. We can also see that no private or public methods are stated for this object, at least in the diagram, and we might examine the class \texttt{Product} for such methods. The database table shows that we can easily begin to design a database for persistent storage of information taken from the story. Being outputs of our thinking, these models provide insights into the nature of our computational-thinking.


And we can, of course, revise the models. We might change the \texttt{condition} attribute to become an enumerated variable, perhaps using the value of \texttt{1} to represent the condition of ``Never worn.'' An enumerated variable would help us implement other features, e.g., with an SQL database, it becomes easier to select products that are never worn, e.g., \texttt{SELECT * WHERE Condition == 1}. An enumerated variable also improves the efficiency of computational processes, e.g., an \texttt{if} test of a numeric variable requires less computational resource than an equivalent test of a string value.

\begin{figure}[ht]
    \centering
    \includegraphics[scale=0.5]{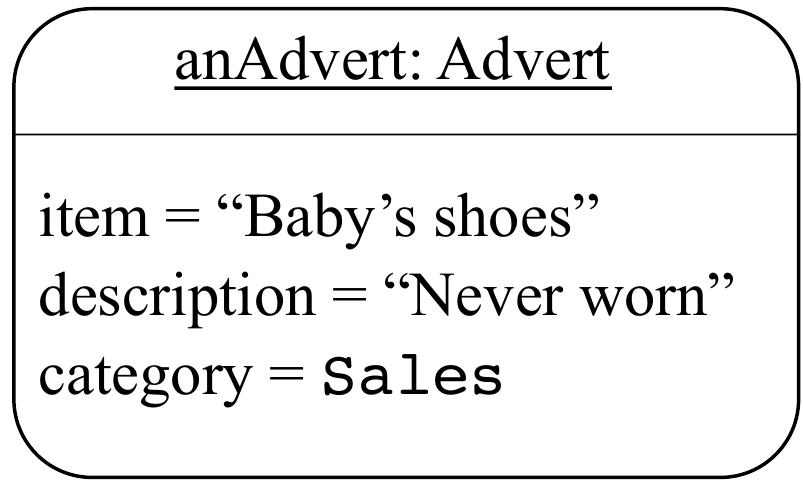}
    \caption{A UML object diagram for an item to advertise}
    \label{figure:advert}
\end{figure}

The object model presented in Figure~\ref{figure:for-sale-diagrams} is not the only model that might be constructed. Figure~\ref{figure:advert} suggests a different object model. Comparing Figure~\ref{figure:for-sale-diagrams} and Figure~\ref{figure:advert}, the same \textit{text} is present but the way we have \textit{structured} the text is different. Our interpretation of the six words leads to different computational representations and those representations then support different thinking, e.g., we might enumerate the \texttt{category} attribute and, with the appropriate database design, select only the sale adverts: the SQL statement, \texttt{SELECT * WHERE category == 1}, now means something different. Though the model has changed, the fundamental nature of the model hasn't, and the kind of thinking -- computational thinking -- hasn't changed either.

\section{Computational-thinking}
\label{section:computational-thinking}

Having presented and discussed examples of outputs from computational thinking, we turn now to consider definitions of computational thinking. We use these definitions to show how computational-thinking is fundamentally different to story-thinking. 

\citeA{aho2012computation} defines computational thinking as, ``\dots the thought processes involved in \textit{formulating problems} so their \textit{solutions} can be \textit{represented as computational steps and algorithms}.'' (emphasis added). \citeA{denning2009profession} recognises the \textit{representation} as more fundamental than the algorithm, since the representation needs to be computable~\cite{erwig2017once} or, on other words, manipulatable by or through computation. \citeA{heinemen2008algorithms} write that, ``Designing efficient algorithms often starts by selecting \textit{the proper data structures} in which to represent the problem to be solved.'' (emphasis added).

\citeA{priami2007computational} highlights a particular feature of the representation: ``\dots the basic feature of computational thinking is abstraction [representation] of reality in such a way that the \textit{neglected details} in the model make it executable by a machine.'' (emphasis added). This concept of negative selection -- i.e., of what is \textit{removed} from the model \textit{in order to} allow or support or enable computation -- is particularly interesting because the focus with abstraction tends to be on what is retained, i.e., on \textit{retaining} only those features of the thing to be modelled that are \textit{essential}~\cite{starfield1994model}. In Section~\ref{section:computational-representations-of-story}, what’s removed from the story is not the literal words but the \textit{order} to the words. Removing the order ``dismantles'' the story. Haven~\citeyear{haven2007story} presents an example to illustrate how simply varying the placement of a word effects the meaning of a sentence. Compare the following examples, taken from Haven's book, \textit{Story Proof}~\cite{haven2007story}:

\begin{displayquote}
    John will marry Elise.\\
    \textit{Even} John will marry Elise.\\
    John will \textit{even} marry Elise.\\
    John will marry \textit{even} Elise.\\
\end{displayquote}

Notice not just that the meaning of the sentence changes but also, if we dwell on each sentence, we might start to wonder about the context and the motivation for John, e.g., what might be going on for John to marry \textit{even} Elise?

As well as ``dismantling'' the story, removing the order of the words so as to make a computable representation also `de-means' -- i.e., reduces the very meaning of -- the story for a human whilst, conversely, formalising the representation to enable computation. At the same time, details are \textit{added} to the computational model, e.g., data types, methods, class-object structures. We return to the issue of ``de-meaning'' later in the paper when we contrast Bruner's Logico-scientific and Narrative modes of thinking. Later in the paper, we also briefly return to the issue of adding information when we discuss software engineering and translation of languages.

Removing information in order to enable computation can be illustrated through another story, the Byzantine Generals Problem (BGP). There are several versions of the BGP reported in the computing literature. We take what is perhaps the most commonly-cited version, published within the safety engineering community by~\citeA{lamport1982byzantine}.

\begin{displayquote}
    \dots several divisions of the Byzantine army are camped outside an enemy city, each division commanded by its own general. The generals can communicate with one another only by messenger. After observing the enemy, \textbf{they must decide upon a common plan of action}. However, some of the generals may be traitors, trying to prevent the loyal generals from reaching agreement. The generals must have an algorithm to guarantee that the following two conditions are met:\\
    \#1: All loyal generals \textbf{decide on the same plan of action} [\dots]\\
    \#2: A small number of traitors cannot cause the loyal generals to adopt a bad plan.\\
    (emboldened emphasises added)
\end{displayquote}

The BGP was formulated by computer scientists to illustrate a particular \textit{computational} problem, i.e., ensuring reliable communication in the presence of faulty components. In a previous publication, we \cite{menon2021stories} critically evaluated the BGP story, concluding that the story `fails' \textit{as a story}. For example, the story lacks any \textit{humanly-meaningful} objective. Generals are expected to agree on a common plan of action, but it does not matter on what they agree. They can attack or retreat or, in principle, do anything else they agree on. Their objective is arbitrary. This is odd for a story because usually in a story a character would have some motive for their objective.
To make a more effective story for the reader (a \textit{reader}, not a software engineer or a programmer or, more generally, a computational-thinker), what is missing from the sentence is a final clause, i.e., decide to do what? Furthermore, since the loyal generals' objective is arbitrary, the traitors' objectives are also arbitrary: the traitors are only concerned with preventing an arbitrary legitimate agreement. But note too that not only must the traitors' objectives be arbitrary relative to the loyal generals, \textit{each} of the traitor's objectives must be arbitrary relative to all other traitors. In this context, each traitor might be better understood as an agent of chaos.

The BGP therefore \textit{risks} misrepresenting the computational problem through the way it presents a story in terms of human agents. To prevent this risk, to ``square the circle'' between story-thinking and computational-thinking, the human agents in the BGP are given odd (for a human) intention.  Safety engineers possess the technical knowledge needed to interpret the story ``correctly'' \textit{for the computational problem} being explored. For safety engineers, as computational-thinkers, the arbitrariness of the objectives is not just acceptable but essential. This is because the arbitrariness of the objectives allows for an algorithmic solution that has wide applicability: the algorithmic solution would apply for situations where generals agree to attack and for situations where generals agree to retreat and for situations where generals (arbitrarily) agree to do something else. The BGP abstracts the problem so as to provide a generalised solution. And in abstracting the problem it must necessarily remove a humanly-meaningful quality of the story, i.e., meaningful human intention.

Overall then, the BGP acts as a kind of mirror example to the six-word story. To go from the six-word story to the computational representation, we remove something essential. To go from the computational problem to a BGP \textit{story} we do not add something essential. In contrasting ways, both neglect. The BGP is not meaning\textit{less} -- we can still understand the story -- but it is not meaning\textit{ful}, in human terms, as a ``good'' story.

\section{Software engineering: programming-in-the-large}
\label{section:programming-in-the-large}

For conciseness, we use Johnson and Ekstedt's~\citeyear{johnson2016tarpit} Tarpit Theory of Software Engineering as our reference for discussing the nature of software engineering. As part of the summary of their theory, Johnson and Ekstedt write, ``The goal of software development is to create programs that, when executed by a computer, result in behavior that is of utility to some stakeholder.''


Drawing on the discussion of computational thinking, we might say that software is \textit{useful} -- of utility -- to a stakeholder when the software behaves in a way that solves a \textit{representation} of a problem experienced by that stakeholder. Solving a representation of a problem is not the same as solving the problem. And inferring from Priami's~\citeyear{priami2007computational} assertion earlier, as well as from our discussion of the BGP, some problems must be \textit{neglected} in order to make the resulting model computable.

Johnson and Ekstedt~\citeyear{johnson2016tarpit} also write that, ``Much of software engineering concerns translations; design specifications are translated into source code\dots, source code is translated into machine code\dots, etc. In fact, the whole process of software engineering can be considered as a series of translations\dots(\citeA{johnson2016tarpit}, p. 187).  They define \textit{language} as ``A set of specifications'', \textit{translation} as, ``An activity that preserves the semantic equivalence between source and target languages [specifications],'' and  \textit{semantic equivalence} as, ``Two specifications are semantically equivalent if, when translated to a common language, they are syntactically identical.'' (Notice how \textit{semantic equivalence} is being defined in terms of \textit{syntactic identity}.)

We have shown, with the six-word story and the BGP story, that it is not necessarily feasible to translate the story from the source language to a target language \textit{and} maintain semantic equivalence.
More than that, it is unreasonable to \textit{expect} a translation, in the way that Johnson and Ekstedt define it, from the English language of the six-word story to, as examples, the User Story, the object diagram or the database table. But that is the point. The language used with story-thinking, and therefore the representations used for story-thinking, are not semantically equivalent to the language and representations used for computational-thinking.

Johnson and Ekstedt~\citeyear{johnson2016tarpit} seem to recognise this problem of non-equivalence, when they write:


\begin{displayquote}
    This leads us to \textit{the major challenge of software engineering}, that programs – these formal, static, syntactic \texttt{compositions} – are very different from the oftentimes \textbf{fluid and intangible stakeholder experiences} they are intended to \textbf{evoke}. A significant feat of \textbf{imagination} is thus required on the part of the \texttt{developers} to predict the effects of a given \texttt{syntactic} modification to the program code on the end-users’s \texttt{experiences}. An even greater challenge, which we propose as the \textit{core task of software} \texttt{endeavors}, is to determine which \texttt{syntactic} manipulation will cause a specified \texttt{stakeholder experience}, and then to perform the appropriate informed manipulations, i.e. to make \texttt{design decisions}. \\
    (\citeA{johnson2016tarpit}; \textit{italicised} emphasis and \texttt{san serif} font are in the original, the \textbf{emboldened} emphasis has been added)
\end{displayquote}

Clearly, Johnson and Ekstedt~\citeyear{johnson2016tarpit} are aware of the ``gap'' between stakeholders' experiences and executable programs, and of the challenge of bridging this gap.  This takes us to requirements engineering, discussed in the next section.

\section{Requirement engineering}
\label{section:requirements-engineering}

In many situations when we use stories, we are not intending to use the stories for requirements gathering or requirements analysis. Requirements engineering has, however, adopted the use of the term ``story'', as well as aspects of stories. Common approaches in requirements engineering are the ``user story''~\cite{lucassen2015forging}, the ``job story''~\cite{lucassen2018jobs}, and the scenario~\cite{carroll2003making, carroll1997scenario}. Requirements engineering (RE) would therefore seem to provide the bridge between stakeholders' ``fluid and intangible experiences''~\cite{johnson2016tarpit}
and the preparation of executable programs. 

But as we showed with the User Stories in Section~\ref{subsection:representation-as-user-story}, this bridge often seems to \textit{begin} the process of neglect, of framing the situation as a \textit{computable} ``problem''. In other words (and however unintentionally), user stories, job stories and scenarios all begin the process of \textit{removing} the non-computable, humanly-meaningful aspects of the situation.

Hinton~\citeyear{hinton2014understanding} provides a good example of this process. He discusses how he and colleagues interviewed clients to gather requirements for a software application that would allow users to change power of attorney for their assets on the website, rather than having to use a paper form.
Interviewees' recollections were about ``getting a form, filling it out, and mailing it in.'' 

Hinton~\citeyear{hinton2014understanding} writes:

\begin{displayquote}
     So, what was missed in these interviews? Most of what was really important, it turns out. Because they [the interviewed clients] knew we were working on a website, these users often unconsciously tried meeting us halfway by \textit{framing} their answers in terms that they assumed we needed. (\citeA{hinton2014understanding}; emphasis added)
\end{displayquote}

and:

\begin{displayquote}
    But if we asked \textit{without priming} the users with that context, their \textit{storytelling} would be closer to the \textit{raw situation} they were in -- for example, ``I just got remarried, and I want to be sure this time my spouse and I have ownership sorted out responsibly,'' and not ``I guess I’d look for the right form to fill out.''~(\cite{hinton2014understanding}; emphasise added)
\end{displayquote}

and also:

\begin{displayquote}
    Meanwhile, the engineers in the company’s IT department wanted to \textit{determine the task} [the problem to be solved] and create a \textit{linear progression} to it. But, we could tell that few people [clients] would recognize that linear path to the task, because the goals that engineering assumed people had already decided on were, in most cases, yet to be discovered. (\cite{hinton2014understanding}, p. 384)
\end{displayquote}

Hinton's account helps to illustrate the two modes of thinking, and the tension arising between them.

\section{Modes of thinking recognised in other disciplines}
\label{section:modes-of-thinking}

Others have recognised differences in modes of thinking, albeit not in the context of programming and software engineering. We briefly consider two complementary perspectives: McGilchrist's two ways of attending, and Bruner's two modes of thought.

\subsection{McGilchrist's two ways of attending}

McGilchrist~\citeyear{mcgilchrist2019master,mcgilchrist2021matter} explores differences in thinking between the left-hemisphere and the right-hemisphere of the brain. To avoid the simplifications of left-brain, right-brain thinking here (which McGilchrist also seeks to avoid ) we'll use the terms ``left-mode attending'' and ``right-mode attending''. These modes of attending can be illustrated through the example of a bird sitting on a branch with a berry in its beak. The bird needs to attend to the berry. This way of attending, the left-mode attending, requires a concentrated, instrumental attention that manipulates the berry-as-thing, a fragment of the world. But at the same time, the bird also needs to attend to the environment. This way of attending, the right-mode attending, requires a broad perspective, in which the bird understands itself in relation to the world as a whole. McGilchrist uses similar examples to explain how these two ways of attending are in opposition to each other: it is not possible for one system to be both narrowly-focused and also, \textit{at the same time}, broadly perceiving. For McGilchrist, this explains, or at least partly explains, an animal's need for two separate hemispheres, where each is a (sub)system that can attend independently.


Drawing on McGilchrist's work, we hypothesise that story-thinking requires right-mode \textit{and} left-mode attending, to both create a story and appreciate it as a reader, but that story-thinking predominantly draws on right-mode attending for its processing and effect. 

\subsection{Bruner's two modes of thought}

Bruner~\citeyear{bruner2020actual} also distinguished two modes of thought, summarised in Table~\ref{table:bruner-modes-of-thought}. The Logico-scientific mode doesn't map cleanly onto computational-thinking, yet there are clear connections with some of the key characteristics identified in the table, e.g., Categorical, General, Abstract, De-contextualised, Non-contradictory, Consistent. Although the Logico-scientific mode of thinking may be self-evidently present, and even dominant, in science and technology, Turner~\citeyear{turner1996literary} argues for the fundamental necessity of the Narrative mode: ``Narrative imagining -- story -- is the fundamental instrument of thought.''

\begin{table}[h!]
\centering
\small
\begin{tabular}{l l l} 
    \hline
    \textbf{Criterion} & \textbf{Logico-scientific} & \textbf{Narrative}  \\ 
    \hline\hline
    Objective   & Truth & Verismilitude \\
    Central problem & To know truth & To endow experience with meaning \\
    Strategy    & Empirical discovery   &   Universal understanding\\    & guided by reasoned hypothesis & grounded in personal experience\\
    Method  & Sound argument    & Good story\\
            & Tight analysis    & Inspiring account \\
            & Reason            & Association \\
            & Aristotelian logic    & Aesthetics\\
            & Proof             & Intuition \\
    Key        & Top-down          & Bottom-up\\
    characteristics        & Theory driven     & Meaning centred\\
            & Categorical       & Experiential\\
            & General           & Particular \\
            & Abstract          & Concrete \\
            & De-contextualised & Context sensitive \\
            & Ahistorical       & Historical \\
            & Non-contradictory & Contradictory \\
            & Consistent        & Paradoxical, ironic\\
    \hline
\end{tabular}
\caption{Jerome Bruner's two modes of thought}
\label{table:bruner-modes-of-thought}
\end{table}

Drawing on Bruner's work, we hypothesise that story-thinking predominantly draws on Narrative thinking for its effect, whilst computational-thinking is predominantly (if not exclusively) a Logico-scientific mode of thinking. 

\section{A summary of the problems and suggested ways forward}
\label{section:the-problem}

Drawing on the preceding discussion, we identify two specific  problems, briefly suggest ways in which these problems might be tackled, and then present candidate hypotheses and research questions for further research.

\subsection{A statement of the problems}

We formulate two problems:

\begin{enumerate}
    \item The problem of neglectful representations. Computational thinking neglects and must neglect. It does this in order to arrive at a representation that is computable. These ``neglectful representations''  inevitably ``de-mean'', i.e., reduce humanly-meaningful qualities. (As a provocative example, consider Turing’s Imitation Game: this reframes intelligence as behaviour, as a \textit{representation} of intelligence, and by doing so ``de-means'' intelligence.) Neglectful representations have implications for the impact of computational-thinking, programming and software engineering on society, the economy and the environment. For example, economic impact is much easier to align with computational-thinking because economic thinking appears to be a way of thinking similar in kind to computational-thinking. But humanly-meaningful qualities of society, as well as the qualities of the environment, are much harder to represent, a point that is increasingly recognised, e.g., with work on values in computing~\cite{ferrario2017values}, kind computing~\cite{alrimawi2022kind}, compassionate computing~\cite{pomputius2020compassionate} and responsible software engineering~\cite{schieferdecker2020responsible}.
    
    \item The opposing differences in the two modes of thinking. The two modes of thinking are not just fundamentally different, but appear opposing, even incompatible. There is then the problem of integrating or otherwise synthesising these two modes of thinking. Achieving some kind of integration would then help to re-balance the first problem.
    
\end{enumerate}

These problems, and the use of story as a possible solution, appear to be implicitly recognised by others in software engineering. \citeA{strom2006reader,strom2007stories} investigates the role and value of stories that include emotions and conflicts in software engineering. Bailin introduced design stories~\cite{bailin2003diagrams} and also discussed how features need stories to convey the ``missing semantics'' that address ``fine-grained questions of context, interface, function, performance, and rationale''~\cite{bailin2009features}. In an unpublished paper, Stubblefield et al.~\citeyear{stubblefield2002micro} observe that, ``In our experience, most software development projects do not fail for technical reasons. They fail because they do not engage users at the fundamental level of value, meaning and practice. They solve the wrong problem, or fail to support deeper patterns of work and social interaction.'' \cite{stubblefield2002micro}.
Finally, Yilmaz et al.~\citeyear{yilmaz2016software} report a preliminary study to demonstrate the benefits of story-driven software development: ``\dots it is important to capture and store the excessively valuable tacit knowledge using a rich story-based approach.'' Significantly, none of these papers consider the cognitive processes that occur with story; in other words, none of the papers consider story in the way we approach it here.

\subsection{Suggested ways forward}

We suggest two directions for future research:
\begin{enumerate}
    

    \item Developing methodology that connects stories with algorithms. In a previous paper, Rainer~\citeyear{rainer2017using}, drawing on prior work in law and legal reasoning, proposed a methodology for identifying arguments and stories from blog articles, extracting them, and then graphically combining them. This methodology might be extended or, alternatively, might provide an example for a potential methodology to integrate representations used for story-thinking with representations used for computational-thinking.
    
    \item Changing software practice. With an agile methodology, software practitioners return to the user story, e.g., its acceptance criteria, to evaluate whether a user story has been delivered. We suggest that practitioners might return to the \textit{story}, and not just the user story. As a further example, software practitioners might initiate one or more reading groups that meet regularly to consider the stories that provide context for the software being developed. For example, a software engineering (SE) team developing a software system for managing information about children in publicly--funded care homes might establish a reading group to read and discuss Sissay's memoir, \emph{My Life is Why}~\cite{sissay2019my}. Or software practitioners who want to ensure their software is being equitable might establish a reading group to read and discuss Ford's memoir, \emph{Think Black}~\cite{ford2021thinkblack}, recounting his father's experience as the first black software engineer at IBM. With these examples, story-thinking and computational-thinking are combined in subtle, informal, but perhaps more culturally effective and longer-term, ways.
    
\end{enumerate}





We also suggest two hypotheses, one assertion and several research questions:

\begin{enumerate}
    \item For the hypotheses:
    \begin{enumerate}
        \item That \textit{programming} requires the kind or kinds of thinking described here as computational-thinking, logico-scientific thinking and left-mode thinking. This is a prescriptive hypothesis, i.e., that programming should be concerned with computational-thinking, logico-scientific thinking and left-mode thinking. In Perlis' words, ``To understand a program you must become both the machine and the program.'' (epigram \#23).
        \item That software engineering is currently oriented toward computational-thinking, logico-scientific thinking and left-mode thinking. Unlike our hypothesis about programming, this hypothesis is normative.
    \end{enumerate}
    
    \item For the assertion:
    \begin{enumerate}
        \item That software engineering \textit{should} comprise a better balance of computational-thinking and story-thinking. Our hypotheses about programming and software engineering are statements of how the world is. By contrast, our assertion is a statement of how the world should be. Software engineering should be better balanced, comprising a balance of both computational-thinking, logico-scientific thinking and left-mode thinking and also of story-thinking, narrative thinking and right-mode thinking.
    \end{enumerate}
    
    \item For the research questions:
    \begin{enumerate}
        \item How do we model humanly-meaningful qualities in software engineering in a way that avoids the problem of neglectful representations?
        \item How do we show that software engineering needs both modes of thinking?
        \item How do we respect and retain both modes of thinking in software engineering practice?
        \item How do we facilitate story-thinking in software engineering?
        \item How do we integrate these oppositional ways of thinking in software engineering?
    \end{enumerate}
\end{enumerate}

\section{Conclusion}
\label{section:conclusion}

In this paper we have explored two modes of thinking, story-thinking and computational-thinking. Using a six-word story as our main example, we show how computational-thinking and story-thinking attend to this story in very different ways. We've considered how software engineering and requirements engineering recognise, at least implicitly, these different modes of thought, as well as the challenges of integrating these modes. We also briefly reviewed two examples of related work, one from  neuroscience and one from psychology. We then identified two specific and fundamental problems (the problem of ``neglectful representations'' and the problem of oppositional ways of thinking), briefly suggested ways in which these problems might be tackled, and proposed hypotheses, an assertion and research questions for future research.

\section{Acknowledgements}
We thank the reviewers in advance for their attention to our paper. We also thank participants of the online presentation for permission to quote them.

\bibliography{references}
\bibliographystyle{apacite} 
\end{document}